# ERGo: Energy-Efficient Hybrid Graph Neural Network Training on Processing-In-Memory Architectures

Pratyush Dhingra, Washington State University, pratyush.dhingra@wsu.edu

Chibuike E. Ugwu, Washington State University, chibuike.ugwu@wsu.edu

Jana Doppa, Washington State University, jana.doppa@wsu.edu

Partha Pratim Pande, Washington State University, pande@wsu.edu

Processing-in-memory (PIM) has been proposed as an alternative computing paradigm for training Deep Neural Networks, including Graph Neural Networks (GNNs). Despite these advancements, training GNN workloads on PIM devices necessitates off-chip memory access. This off-chip access is expensive in terms of performance and energy, thereby impacting the overall energy efficiency of the training process on PIM platforms. In this paper, we propose a novel hybrid training framework called ERGo that automatically switches from a full-parameter phase to a parameter-efficient phase during the training process with negligible loss in predictive accuracy. The parameter-efficient phase employs low-rank representations of GNN weights, effectively reducing the number of trainable parameters. This helps in reducing the off-chip access during the end-to-end training process on PIM architectures, leading to notable improvements in both performance and energy efficiency. ERGo outperforms the conventional full-parameter GNN training on a PIM-based platform by up to 3.15x in speedup and enhances energy efficiency by 10.5x. Furthermore, ERGo offers additional advantages when utilized on heterogeneous architectures incorporating non-volatile memory (NVM) and static random-access memory (SRAM). Training on the heterogeneous NVM-SRAM architecture typically utilizes NVM for the forward-pass and SRAM for the backward-pass. However, NVM cells suffer from low device endurance and repeated write operations due to weight updates can lead to poor lifetime. ERGo improves the lifetime of NVM devices by freezing the weights and preventing weight updates in the parameter-efficient phase. Experimental results demonstrate that ERGo improves the lifetime of heterogeneous NVM-SRAM architecture by up to 33x in comparison to full-parameter implementation.

CCS CONCEPTS • Hardware → Emerging technologies • Computing methodologies → Machine learning • Hardware →Very large scale integration design → On-chip resource management

**Additional Keywords and Phrases:** PIM, GNNs, Heterogeneity, Reliability, Non-Volatile Memory, Resource-aware ML

## 1 INTRODUCTION

Graph Neural Networks (GNNs) are a class of deep learning models designed to operate on graph-structured data. GNNs are utilized extensively across diverse application domains, including chemistry, biology, social networking, recommender systems, and computer vision [1]. The training of GNNs consists of two key computational kernels: *aggregation* and *combination*, which require both graph data and trainable parameters, as shown in Fig. 1 [2]. As a result, GNN training is both memory- and compute-intensive. This makes the achievable performance of conventional computing platforms, such as CPUs/GPUs, limited during training with the processing cores waiting for off-chip data. The limited memory bandwidth (memory wall) and low compute efficiency (FLOPs/W) have necessitated the exploration of energy-efficient domain-specific accelerators for GNN training.

Processing-in-memory (PIM)-enabled architectures have garnered strong interest as an alternate computing paradigm to enable high-performance and energy-efficient Deep Neural Network (DNN) training and inferencing [3]. PIM platforms achieve superior performance and energy efficiency by minimizing costly memory access for computation. In this regard, several non-volatile memory (NVM)-based PIM accelerators have been proposed for GNN training [4]. However, the back-propagation computation during GNN training requires multiple write operations, and NVM cells suffer from limited write endurance and high write latency/energy [5]. Alternatively, Static Random-Access Memory (SRAM) based PIM accelerators have also been proposed for DNN training. SRAM cells offer low write operation latency and do not suffer from low device endurance [6]. Nevertheless, SRAM-based PIM architectures are associated with significant area overhead and high leakage energy compared to their NVM counterparts. Consequently, heterogeneous integration has been proposed to address the limitation of homogeneous PIM accelerators. Specifically, SRAM has been used for the backpropagation phase, which requires multiple write operations and high-precision computation, while NVM is utilized to implement the forward-propagation computation [7]. Despite these advancements, the memory demands of training modern deep learning workloads often necessitate off-chip memory access on PIM devices. The on-chip buffers are mostly insufficient to store the large amount of intermediate data required for weight updates. This results in off-chip access, which is expensive in terms of performance and energy, thereby impacting the overall energy efficiency of the training process on both homogeneous and heterogeneous PIM platforms.

Recently, low-rank representations have been explored for fine-tuning foundation models for downstream predictive tasks [8]. Low-Rank Adaptation (LoRA) fine-tunes the base model by updating a fraction of the parameters. In this context, LoRA matrices serve as trainable parameters while maintaining the weights of the pre-trained model fixed [9]. This approach significantly reduces the computing and memory requirements during the back-propagation computation. Thus, exploring LoRA to address the limitations associated with GNN training on PIM architectures is a promising direction. However, it is important to note that LoRA has primarily been applied for fine-tuning tasks on transformer models [9]. We need to address some critical challenges to use it for training GNNs from scratch.

In this work, we introduce a hybrid framework for GNN training on PIM architectures called ERGo. The training of GNNs, as represented by the model accuracy over training epochs, does not follow a linear trajectory. This non-linear characteristic (the rate of improvement in training accuracy is relatively small after a few initial training epochs) presents an opportunity to adapt the training strategy, thereby enhancing both the performance and energy efficiency of the hardware platform. The critical step in our approach involves transitioning to a parameter-efficient phase utilizing LoRA after the full-parameter phase during the first few training epochs (see illustration in Fig. 1). To maintain accuracy and reliability during the shift from the full-parameter phase to the parameter-efficient phase, we need to automatically determine the training epoch for this transition in a reliable manner. To address this challenge, we adopt an algorithm with provable guarantees based on the framework of conformal prediction (CP) which allows us to determine the transition epoch in a reliable manner [10] [11]. We train a regression model on augmented data using the characteristics of the GNN training instances and calibrate it using CP to quantify the uncertainty over the switching epoch in the form of a lower- and upper-bound. CP guarantees that the true switching epoch is within these two values with a high probability, say 95% (parameter value used to configure the conformal calibration process). Therefore, selecting the upper bound as the switching epoch in ERGo enhances the overall efficiency of the training without loss in predictive training accuracy [12]. This hybrid approach enables reliable, energy-efficient, and high-performance GNN training. The key contributions of this paper are as follows:

1. We propose a novel hybrid training methodology, ERGo, designed to efficiently train GNNs on PIM architectures. ERGo employs a parameter-efficient phase during training to enhance energy efficiency and performance across various GNN models and graph datasets. We develop a theoretically grounded conformal prediction approach to automatically select the transition epoch to switch from full-parameter to parameter-efficient training to achieve a targeted trade-off between accuracy, performance, and energy-efficiency.
2. We show that ERGo is applicable to existing homogeneous and heterogeneous PIM accelerators for GNN training. Notably, ERGo enables superior performance and energy-efficiency while maintaining comparable accuracy to that of full-parameter training (less than 1% test accuracy loss).
3. We demonstrate that ERGo, unlike existing state-of-the-art techniques for lifetime improvement, extends the lifetime of NVM cells in a heterogeneous NVM-SRAM PIM accelerator without introducing performance and energy overhead.

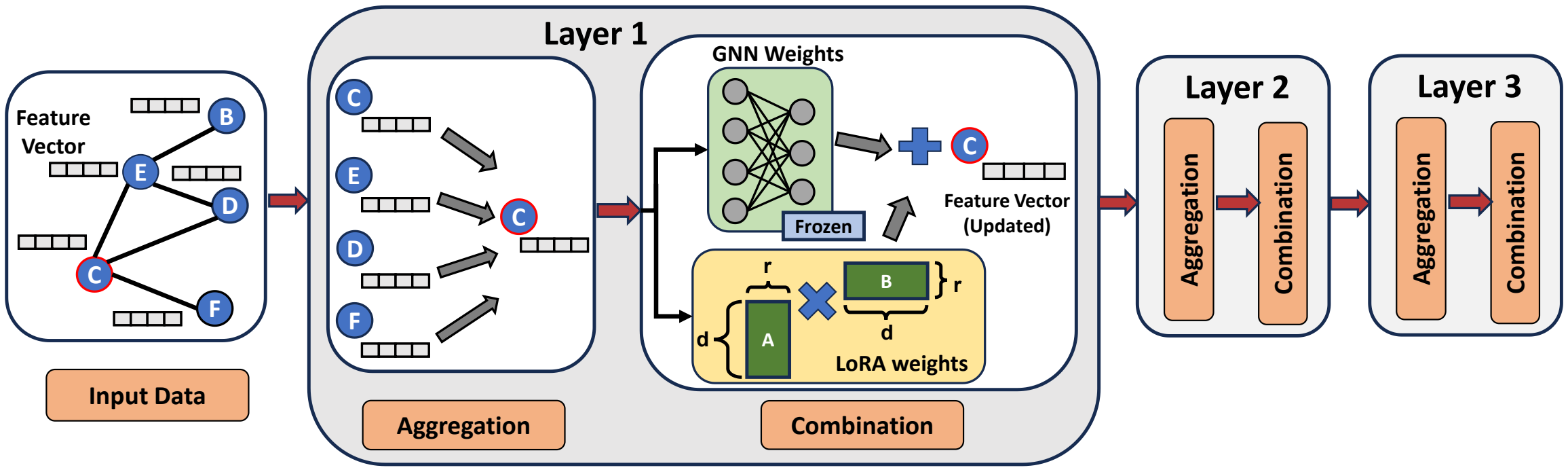

Figure 1: The GNN computational kernel showing aggregation and combination for the highlighted node C as an example during the parameter-efficient phase in ERGo.

*To the best of our knowledge, this is the first work that proposes a hybrid training framework for training GNNs from scratch*. The rest of this paper is organized as follows; Section 2 discusses prior work; Section 3 outlines the GNN training computation kernel, and Section 4 elaborates on the ERGo framework. Section 5 presents the experimental results, and Section 6 concludes the paper by summarizing the key findings of this work.

## 2 RELATED WORK

In this section, we present relevant existing work on GNN accelerators and discuss the endurance issue in NVM devices.

### 2.1 GNN Training Accelerators

Various Von Neumann architecture-based accelerators have been developed to support GNN training, including HyGCN, FlowGNN, and NEM-GNN [13] [14] [15]. However, they are constrained by memory bandwidth limitations. To solve the memory bottleneck, a broad array of PIM-based accelerators has been proposed for GNNs. NVM-based PIM accelerators proposed include PIMGCN, ReFLIP, Re-GNN, Fe-GCN, ReMaGN, and DARe [4] [16] [17] [18] [19] [20]. ReFLIP, ReMAGN, and DARe are ReRAM-based PIM accelerators that leverage Ohm's and Kirchhoff's current laws to perform $N^2$ matrix-vector multiplications in $O(1)$ time in the analog domain, making it extremely energy efficient and achieving high performance [4] [19] [20]. Fe-GCN adopts FeFET instead of ReRAM as the NVM device for GNN training [16]. In contrast, ReGNN is a ReRAM-based GNN accelerator that performs computation in both digital and analog domains. It utilizes analog computing for accelerating matrix-vector multiplications (MVMs), while non-MVM operations are performed in the digital domain [17]. However, the GNN training necessitates weight update after processing each batch. This back-propagation computation and weight update during the training process leads to repeated writes and result in reliability concerns for NVM cells [21]. Furthermore, device variations during the back-propagation computation can result in degraded model accuracy [6]. Digital SRAM-based PIM accelerators have been proposed for GNNs to address the endurance concerns of NVM-based designs. DCIM-GCN, for example, incorporates a sparsity-aware compute engine tailored for efficient aggregation in GNNs [22]. Nevertheless, it is limited to small graph datasets due to the large area footprint of SRAM cells relative to NVM-based counterparts [23]. HD-CIM proposes a heterogeneous memory architecture, using NVM for storing model weights and SRAM for computation [24]. However, HD-CIM architecture is more suitable for inference tasks given the large memory requirement for storing intermediate activations during training. Heterogeneous architectures featuring NVM-SRAM have also been explored for high-accuracy DNN training [25]. Specifically, SRAM is utilized for the back-propagation computation in the digital domain to avoid issues related to device variations [25]. Additionally, SRAM PIM offers high endurance and low access speed in comparison to its NVM counterpart, making it suitable for backpropagation computation [26]. PIM accelerators adopt mini-batch training, which generates activations that must be stored for subsequent weight updates. These activations and errors may need to be stored off-chip given the limited on-chip memory. Consequently, the frequent off-chip access during back-propagation computation reduces the overall energy efficiency of the training process on the PIM platform. In this work, we reduce the off-chip storage demands by using a parameter-efficient phase during GNN training. DRAM-based PIM architectures have also been proposed for DNNs. However, these architectures are primarily employed for inference tasks. In contrast to inference, NN training is more complex, as it involves gradient computation during back-propagation. Further, the 1T1C structure of DRAM cells lacks inherent in-situ computation capabilities, which limits its ability to support parallel MVM operations [27]. This parallelism is necessary for achieving high-throughput energy-efficient DNN training.

### 2.2 Lifetime Improvement in NVM-based PIM Accelerators

The write endurance of NVM devices typically ranges from $10^6$ to $10^{12}$ writes. Consequently, frequent writes to NVM cells can lead to device failure [26]. However, the weight updates necessary for DNN training would require repeated writes, thereby significantly reducing the overall lifetime of NVM devices. In this context, several methods have been proposed to mitigate the effect of low write endurance of NVM devices, encompassing both hardware and software-based strategies. Hardware approaches do not reduce the number of writes but add redundancy to the computing system, such as redundant columns or a fault map to compensate for the output [28]. These approaches incur significant area and energy costs for on-device training. Conversely, software approaches involve algorithmic techniques for lifetime improvement. A popular technique is read-before-write, which reads the current bit values of a memory location and updates only those bits that are different, thereby reducing the number of rewrites [29]. A similar approach is bit flipping, where the bits to be written are compared with the new data and mapped either in the original format or inverted format to minimize the number of rewrites [30] [31]. Another technique to improve lifetime is row/column swapping, which aims to distribute writes more evenly across a crossbar [32]. Specifically, the rows/columns that are frequently written are periodically swapped with the ones updated with a lower frequency. However, the performance overhead of repeated remapping is significant as it leads to stalls during the pipelined implementation [33]. Structured gradient sparsification (SGS) has also been proposed to mitigate low device endurance [34] [35]. This approach reduces the number of writes by value-based gradient pruning. Similar to remapping, SGS increases the number of on-chip activations as gradients after pruning are accumulated over multiple batches. In this work, we reduce the number of write operations to the NVM crossbars without introducing performance overhead using an algorithmic approach.

## 3 GNN TRAINING

In this section, we discuss the GNN computational kernel and provide a brief background on low-rank decomposition.

### 3.1 GNN Computational Kernel

The GNN computational kernel for each layer is split into two phases: *aggregation* and *combination* [1]. The aggregation phase computes the aggregated node features from neighboring nodes using the graph adjacency matrix. Conversely, the combination phase transforms the aggregated feature vector using learnable weights to generate the updated feature vector for the subsequent layer [1]. Eq. 1 and 2 describe the aggregation and combination phase computational kernels, respectively for a node $v$ :

$$GNN\ layer\ l: \begin{cases} y_v^l = Agg(f_v^{l-1} \times Adj_i) & (1) \\ f_v^l = Comb(W^l \times y_v^l) & (2) \end{cases}$$

Here, $f_v^{l-1}$ denotes the input feature vector from the previous layer, $Adj_i$ represent the adjacency matrix, and $y_v^l$ represents the aggregated node features at layer $l$. The updated feature vector $(f_v^l)$ is obtained by multiplying the weight matrix $W^l$ with $y_v^l$. The $Agg$ and $Comb$ represents the aggregation and combination function specific to each GNN model [2]. Both these kernels can be reduced to MVM operations, which can be effectively implemented on PIM devices [36].

### 3.2 Low-Rank Adaptation

Existing works have shown that pre-trained representations have an intrinsic dimension several orders of magnitude less than that of full parameterization [37]. LoRA is a fine-tuning technique that builds upon this concept by introducing trainable low-rank matrices, keeping the parameters of the original base model unchanged [9]. Specifically, the model weights for a layer ($W \in R^{d\times d}$) are decomposed into a sum of frozen matrix ($W_O \in R^{d\times d}$) and a low-rank matrix ($AB$) :

$$W = W_O + AB \qquad (3)$$

Here, $A \in R^{d\times r}$ and $B \in R^{r\times d}$ are the two trainable low-rank matrices where "$r$" is the rank and "$d$" represents the dimensionality of the original weight matrix. The trainable low-rank matrices are considerably smaller in size compared to the original weight matrix with $r \ll d$ [9]. For instance, prior work has shown that the value of $r$ in the range of 8 to 32 is sufficient for a model with $d = 2048$ [9]. This reduces the computational complexity of the trainable parameters from $O(d^2)$ to $O(d.r)$ for a given layer. Fig.1 shows low-rank adapters added to the weight matrices in a GNN layer. Note that LoRA can only be applied to trainable parameters, and the aggregation operation involves multiplication with the graph adjacency matrix which is not trainable.

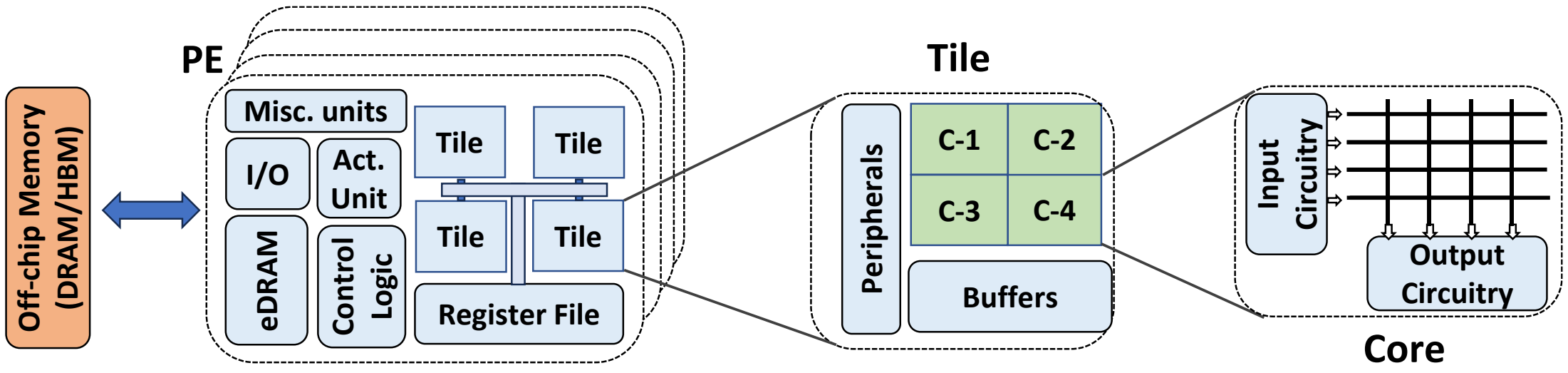

Figure 2: A conceptual illustration of the PIM processing element and memory hierarchy.

## 4 ERGO: A HYBRID GNN TRAINING FRAMEWORK

In this section, we introduce the hybrid training framework, ERGo. First, we discuss the GNN training on PIM architectures. Subsequently, we discuss the parameter-efficient phase in ERGo, which utilizes LoRA. Next, we describe the conformal prediction algorithm that determines when to switch to the parameter-efficient phase during GNN training. Finally, we provide details of the end-to-end training process with ERGo.

### 4.1 GNN Training in PIM Architectures

PIM architectures leverage pipelined implementation with mini-batch training to achieve high throughput [21]. In the context of GNNs, the input monolithic graph $(G)$ is initially partitioned into smaller subgraphs during pre-processing, expressed as $G = \sum_{j=1}^{p} SG_j$ . The number of graph partitions $(p)$ is determined considering the on-chip resource constraints [4]. These subgraphs are then processed in batches, with each subgraph $(SG_j)$ sent sequentially as input into the pipeline [4]. The pipelined implementation necessitates the execution of all GNN layers simultaneously. Subsequently, the weights of all GNN layers and the adjacency matrix for the batch of subgraphs under consideration are mapped onto the PIM crossbars. Note that the adjacency matrix, which represents graph data, remains the same across all the layers. During training, back-propagation requires the activations generated during the forward propagation for gradient computation. Given a batch size of $B$ subgraphs, the forward-propagation generates $B$ activations per layer. These activations are utilized for computing the error during back-propagation. Subsequently, the gradient computation necessitates access to the $B$ activations and the corresponding $B$ errors across all layers. Once the gradients are computed, the GNN weights are updated on the PIM crossbars. Additionally, the adjacency matrix for the next batch of subgraphs is also updated on the crossbars.

Fig. 2 presents a conceptual processing element (PE) and the memory hierarchy in PIM architectures. Each tile contains multiple cores with crossbars that store weights and adjacency matrices. During training, the input subgraphs are loaded from DRAM to the on-chip memory in each PE, where the control circuitry sends data to input buffers in each core to perform MVM operations. The resulting output activations are stored in the on-chip buffers/eDRAM and serve as input activations for subsequent layers. However, the intermediate activations required for the back-propagation computation need significant on-chip memory. It may be infeasible to store all activations on-chip due to area and power limitations. In scenarios where intermediate activations exceed the on-chip storage capacity, they are stored off-chip in DRAM.

Homogeneous NVM-based PIM architectures, including ReMaGN, ReFLIP, and ReGNN [4] [17] [19], carry out both forward and back-propagation on NVM crossbars. This reduces the device lifetime since the gradient computation needs multiple writes to compute errors for each subgraph in the batch. Therefore, we consider commonly used heterogeneous NVM-SRAM PIM architecture as the hardware platform for GNN training [25] [23]. The use of SRAM PIM can mitigate the endurance issues during back-propagation computation, as SRAM does not suffer from limited write endurance [26].

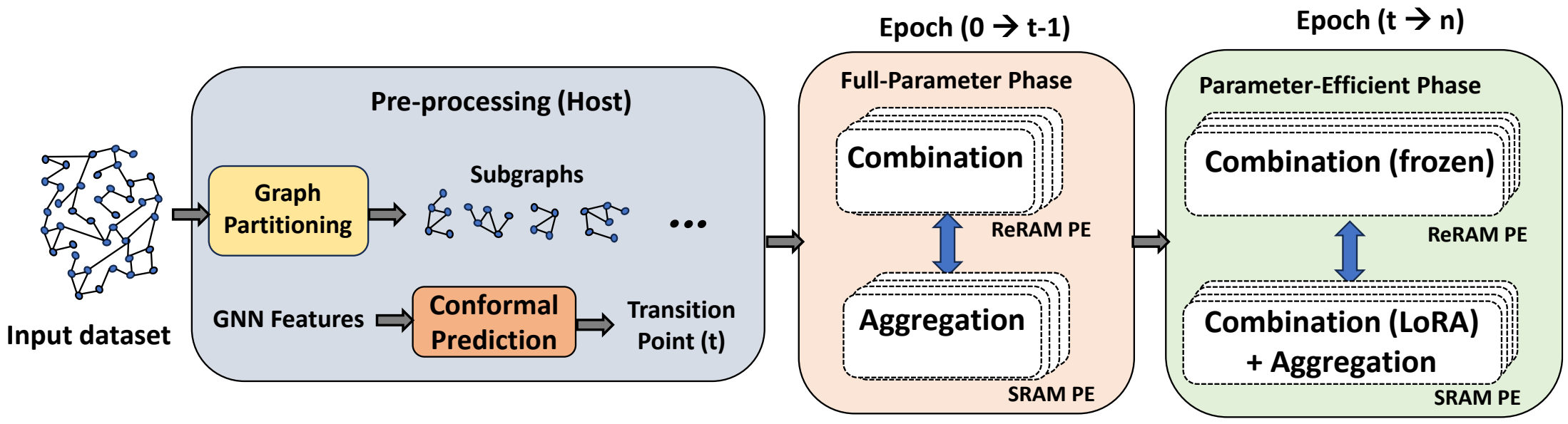

Figure 3: High-level overview of the end-to-end GNN training process with ERGo.

However, even with SRAM integration, periodic writes remain necessary after processing each batch to update weights and adjacency matrices stored in NVM crossbars. Moreover, this approach does not alleviate the requirement for off-chip DRAM access during training. We try to address the concerns regarding GNN training on PIM platforms.

### 4.2 ERGo Training Framework

In this work, we propose ERGo, a novel training framework that incorporates two distinct phases: a full-parameter phase and a parameter-efficient phase. We employ full-parameter phase during the initial few epochs (e.g., 5-10 percent of total epochs) of the training process. The full-parameter phase utilizes NVM for forward-pass computations due to its high density and low static power consumption, whereas SRAM is utilized for backward-pass operations because of its superior write endurance and speed characteristics [23]. Additionally, SRAM is utilized instead of NVM for the aggregation phase. Following the initial full-parameter phase, the training process transitions to the parameter-efficient phase, as illustrated in Fig. 3. The transition point (switching epoch) is determined prior to the training phase in the pre-processing step using an algorithmic approach based on conformal prediction (CP). CP allows us to determine the transition epoch in a reliable manner with a high probability, say 95% (parameter value used to configure the conformal calibration process). We detail our algorithm based on the CP framework in subsection 4.4.

In the parameter-efficient phase, the GNN weights associated with the full-parameter phase are frozen and remain on-chip in NVM PEs. We initialize additional trainable LoRA parameters to represent these frozen weights. The LoRA parameters are trained using backpropagation during the parameter-efficient phase and mapped on SRAM PEs due to their superior endurance characteristics. The aggregation phase remains unchanged since it involves multiplication with the adjacency matrix. Fig. 3 provides a high-level overview of the end-to-end hybrid training framework, ERGo. Note that utilizing a homogeneous SRAM architecture for traditional full-parameter GNN training instead of ERGo would have required significantly higher SRAM resources. For instance, utilizing LoRA with a rank of 32 for weights with dimension $1024 \times 1024$ results in a 16-fold reduction in SRAM resource requirements compared to storing the weights with the original dimension in SRAM PEs. Moreover, the use of low-rank matrices reduces both the gradient computation as well as the storage requirement of gradients during the backpropagation computation.

### 4.3 Parameter-Efficient Phase

We now detail the computation kernels in the parameter-efficient phase. The GNN forward-propagation with the introduction of LoRA changes as follows:

$$GNN\ layer\ l: \begin{cases} y_v^l = Agg^l(f_v^{l-1} \times Adj_i) & (4) \\ f_v^l = Comb^l((W_0{}^l + A^l B^l) \times y_v^l) & (5) \end{cases}$$

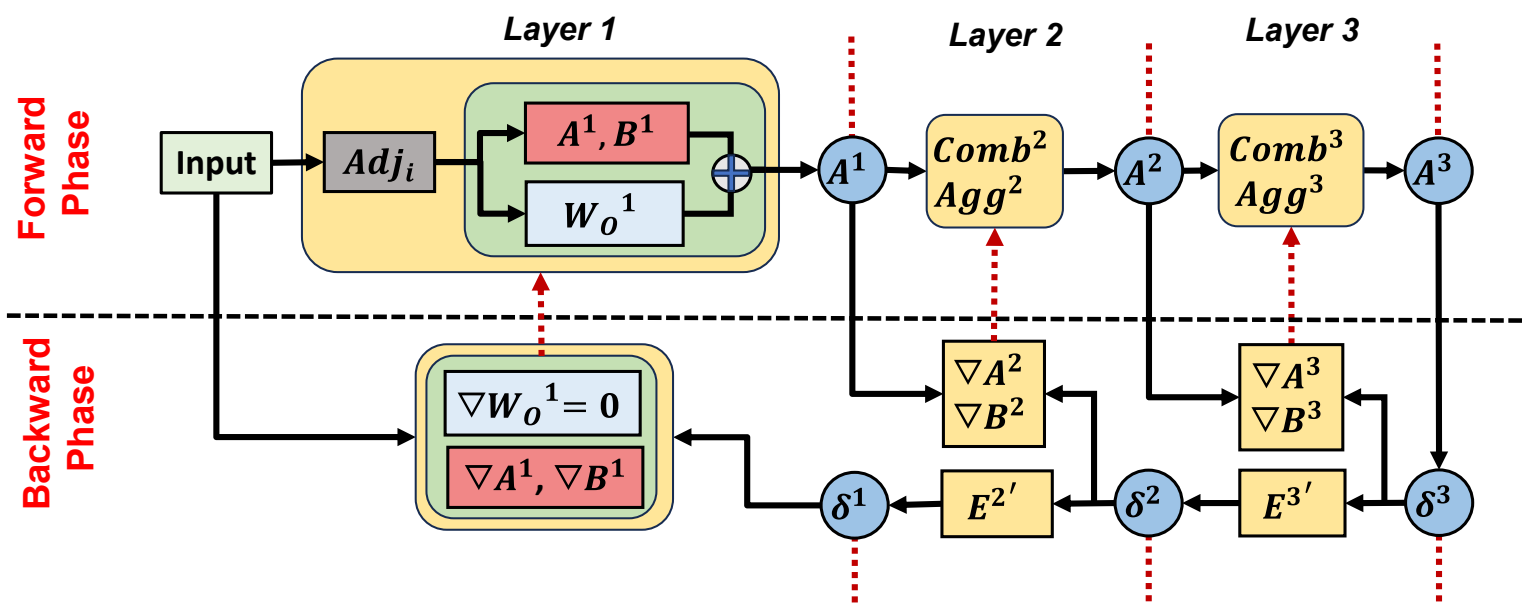

Figure 4: Parameter-efficient phase training pipeline presenting both forward and back-propagation computation. The gradients are generated only for the LoRA matrices A and B, while the gradients for frozen matrices are zero.

The aggregated feature vector ($y_v^l$), which serves as input for the combination phase, is sent to the NVM crossbars where the frozen weights ($W_O{}^l$) are stored as well as the SRAM blocks where low-rank matrices ($A^l$ and $B^l$) are mapped. The output activation is obtained by the summation of the MVM operations involved in both computations. Fig. 4 details the forward-propagation mapping.

Now, we determine the computational complexity of gradient computation during the parameter-efficient phase of training. The back-propagation step computes the gradient of the loss with respect to the weight matrix. Let $\delta^l$ be the error for the layer $l$. Considering $W$ as the weight matrix during the full-parameter phase, the gradient of loss $L$ with respect to $W$ for layer $l$ can be expressed as:

$$\nabla W^l = \frac{\partial L}{\partial W} = \delta^l . (f^{l-1})^\top \tag{6}$$

During the parameter-efficient phase, we need to calculate the gradient of the loss $L$ with respect to $A^l$ and $B^l$, since these two are the only trainable parameters in a layer. The gradients for the frozen matrix ($W_O{}^l$) are zero.

$$\nabla A^l = \frac{\partial L}{\partial A^l} = \delta^l . (f^{l-1})^\top . (B^l)^\top \tag{7}$$

$$\nabla B^l = \frac{\partial L}{\partial B^l} = (A^l)^\top . \delta^l . (f^{l-1})^\top \tag{8}$$

Here, the gradients $\nabla A^l$ and $\nabla B^l$ are of size $d \times r$ and $r \times d$, respectively, which are much smaller than the full gradient $\nabla W^l$ of size $d \times d$. This reduction in gradient size decreases the amount of off-chip data access to DRAM. Additionally, this computational mapping prevents further write operations to NVM cells associated with weight updates since only the LoRA matrices are updated during back-propagation, which utilizes SRAM PIM. Fig. 4 shows the parameter-efficient GNN training pipeline, highlighting the forward and backward propagation.

Overall, the advantages of the parameter-efficient phase are two-fold. First, we significantly reduce expensive off-chip data access during the back-propagation computation, thereby improving energy efficiency. Importantly, the reduction in off-chip access is independent of the specific type of PIM architecture employed. ERGo helps enhance the overall performance and energy efficiency of the system regardless of the underlying PIM architecture. We later show in the experimental section that the end-to-end hybrid training framework achieves higher energy efficiency across various PIM hardware platforms, including existing homogeneous NVM-based GNN accelerators. Second, adopting ERGo on a

ALGORITHM 1: Creation of Calibrated Regression Model via Conformal Prediction

**Input**: $\{gnn$ = GNN model, $G$ = Graph dataset$\}$ = set of GNN Training workloads, $f$ = regressor, $\alpha$ = user-defined error rate
**Output**: $C$ = conformal calibrated regression model with coverage $1 - \alpha$ // $(1 - \alpha)$ = 95% in our-case

*/ Creation of augmented GNN training instances and supervised data for regression */
Set of augmented GNN training instances, $I \leftarrow \phi$
Training examples for regression task, $D \leftarrow \phi$
**for** each given GNN training workload $(gnn, G)$ **do**
    $\sum_{j=1}^{p}(SG_j) \leftarrow G$ // Partition G into $p$ subgraphs
    Vary the number of subgraphs and the number of node features : $\{gnn_k, \sum_{j=1}^{k}(SG_k)\}$
    $I \leftarrow I \cup \{gnn_k, \sum_{j=1}^{k}(SG_k)\}$ // Add augmented GNN training instance to I
**end for**
**for** each augmented GNN training task $gnn_i \in I$ **do**
    Experimentally identify the true transition point $(Y_i)$ such that acc. loss is within user-defined threshold
    $D \leftarrow D \cup (gnn_i, Y_i)$ // Add true transition point corresponding to each task in D
**end for**
Divide D into training $(\mathcal{D}_{tr})$ and calibration $(\mathcal{D}_{cal})$ sets such that $\mathcal{D}_{tr} \cap \mathcal{D}_{cal} = \phi$
Train $f$ using the training data $\mathcal{D}_{tr}$
Calibrate $f$ using CP with $1 - \alpha$ coverage on the calibration data $\mathcal{D}_{cal}$ to get calibrated regression model $C$
**Return** conformal calibrated regression model with coverage $1 - \alpha$, $C$

heterogeneous NVM-SRAM PIM platform for GNN training offers a distinct advantage: it extends the lifetime of NVM devices by eliminating rewrites to NVM cells during the parameter-efficient phase of training.

The key remaining challenge is to determine the appropriate timing for transitioning from the full-parameter phase to the parameter-efficient phase. The epoch at which this transition occurs has a direct impact on the overall efficiency of the training process and the model accuracy. A delayed transition may result in missed opportunities for energy savings, while an early transition could compromise the accuracy of the trained GNN model.

### 4.4 Conformal Prediction Algorithm to Determine Transition Point

We need to determine the transition point from the full-parameter phase to the parameter-efficient phase in ERGo. In this work, we define the transition point as the percentage of the minimum number of epochs in the full-parameter phase, required to preserve model accuracy within a user-specified margin (say less than 1% loss). It can be expressed as:

$$transition\ point = \frac{epochs_{full-parameter}}{epochs_{total}} \times 100 \tag{9}$$

The identification of the transition point consists of two key steps. First, we train a regression model $(f)$ to predict this transition point for a given GNN dataset based on relevant features. Second, we employ conformal prediction (CP) to calibrate this regression model to obtain uncertainty estimates in the form of lower-bound and upper-bound such that the true transition point lies within them with a high probability $1 - \alpha$ (say 95%) which is specified as a parameter for the calibration process. For an unseen GNN dataset, we use the calibrated regression model to get the bounds and select the upper-bound as the transition point $(t)$ as part of the pre-processing step on the host device, as illustrated in Fig. 3. The algorithmic steps, namely, training and calibration of the regression model require supervised training data. The three key steps of our overall methodology: creation of supervised training data, training of the regression model, and calibration of the regression model, are performed offline as a one-time approach. Algorithm 1 details the creation of a calibrated regression model to determine the transition point. Below we provide the details of these three key steps.

**Creation of augmented GNN training instances -** We create supervised data for these two steps using a given set of GNN training instances as follows. First, we create diverse *augmented* GNN training instances using the following transformations: select a subset of subgraphs (less than the total subgraphs) and/or select a subset of node features (less than the total features). Second, for each augmented GNN instance (input), we run full-parameter GNN training to benchmark accuracy and determine the earliest transition point (output) to preserve accuracy, e.g., less than 1% loss in accuracy. This supervised data, in the form of GNN training instances and the corresponding true transition point pairs, is used to train and calibrate a regression model to determine the transition point for unseen GNN datasets.

**Training regressor to predict the transition point.** We train a Random Forest quantile regressor following the existing work [38], using the following features of augmented GNN training instances: the no. of features of each node, the no. of subgraphs, the no. of edges, and the number of output labels to predict. Note that we can utilize any other regression model or feature set in our overall methodology.

**Conformal prediction for calibration -** The regression model may exhibit potential inaccuracies in predicting the transition point when applied to unseen GNN datasets. This can either lead to a loss in GNN model accuracy (earlier than the true transition point) or a loss in performance/energy efficiency (later than the true transition point). Hence, our goal is to preserve accuracy and minimize loss in performance/energy efficiency. Motivated by this goal, we employ the CP framework for calibrating the trained regression model $f$. CP uses calibration data to configure the calibration process for a user-specified value of coverage $1-\alpha$ (we use 95%). After calibration of the regression model, for a new unseen GNN dataset $X_{test}$, we get a lower-bound ($lb$) and upper-bound ($ub$) for the transition point, i.e., uncertainty set $C(X_{test}) = [lb, ub]$. CP provides a theoretical guarantee that the true transition point $Y_{test}$ corresponding to the GNN training task $X_{test}$ lies within the uncertainty set $C(X_{test})$ with a very high probability $1-\alpha$ [38] [12].

$$\mathbb{p}[Y_{test} \in C(X_{test})] \geq 1-\alpha \quad (10)$$

This means that, with high probability (95% in our case), the true transition point is guaranteed to fall within $lb$ and $ub$. To ensure that we balance both accuracy and the required coverage level during transition point selection, we choose $ub$ as the transition point for ERGo. We now present theoretical proof that this choice guarantees that even in the worst-case scenario, our selection is still correct with at least $1-\alpha$ probability.

Let $(X_i, Y_i)$ denote any augmented training example created for calibration purpose in Algorithm 1: $X_i$ corresponds to the features of a GNN training task and $Y_i$ corresponds to the true transition epoch determined experimentally.

**Theorem 1.** *Given that $(X_i, Y_i), i = 1,2, \ldots, n+1$ are exchangeable, i.e., the order in which they appear does not affect the joint probability distribution. Let $C(x) = [lb, ub]$ denote the prediction interval produced by the conformal prediction algorithm, constructed using* n *datapoints. Then, for a new test data point $X_{n+1}$ (i.e., unseen GNN training task) and any user-defined error rate $\alpha \in (0,1)$, the $C(X_{n+1})$ set satisfies the following coverage guarantee:*

$$\mathbb{p}[Y_{n+1} \in C(X_{n+1})] \geq 1-\alpha \quad (11)$$

**Proof:** Given any base quantile regression algorithm, we begin by fitting the upper and lower quantile functions $q_{\alpha 1}$ *and* $q_{\alpha 2}$. Let $S_i$ be the conformity score that measures the degree of discrepancy between the predicted value and the ground truth on each calibration example from the set $D_{cal}$, such that:

$$S_i \coloneqq \max\{q_{\alpha 1}(X_i) - Y_i, Y_i - q_{\alpha 1}(X_i)\} \forall i \in D_{cal} \quad (12)$$

For a new test point $(X_{n+1}, Y_{n+1})$, by the construction of the prediction interval, we get:

$$Y_{n+1} \epsilon\ C(X_{n+1})\ if\ and\ only\ if\ E_{n+1} \leq\ Q_{1-\alpha}(E, D_{cal}) \tag{13}$$

Where $Q_{1-\alpha}(E, D_{cal})$ is the $(1-\alpha)(1+\frac{1}{|D_{cal}|}|)$-th empirical quantile of $\{E_i: i \in\ D_{cal}\}$

Particularly,

$$\mathbb{p}\{Y_{n+1} \epsilon\ C(X_{n+1})|(X_i, Y_i), i = 1, \dots, n\}\ =\ \mathbb{p}\{E_{n+1} \leq\ Q_{1-\alpha}(E, D_{cal})|(X_i, Y_i), i = 1, \dots, n\} \tag{14}$$

Since the pairs $(X_i, Y_i)$ are exchangeable, then the calibration conformity scores $E_i$ are also exchangeable for $i \in\ D_{cal}$ and $i = n+1$. Therefore,

$$\mathbb{p}\{E_{n+1} \leq\ Q_{1-\alpha}(E, D_{cal})|(X_i, Y_i), i = 1,2, \dots, n\} \geq 1-\alpha \tag{15}$$

$$\Rightarrow\ \mathbb{p}\{\boldsymbol{Y_{n+1}\ \epsilon\ C(X_{n+1})}\} \geq \boldsymbol{1-\alpha} \tag{16}$$

This proves Theorem 1.

**Corollary:** *Let $C(x) = [lb, ub]$ be a conformal prediction interval with coverage level of at least $1-\alpha$. This means that, with probability at $1-\alpha$, the true value Y lies somewhere within the interval $[lb(x), ub(x)]$. Then, if we choose the upper bound $ub(x)$ as our decision point (i.e., the training epoch at which we switch from full training to parameter-efficient training in our setting), we are guaranteed—again with probability at least $1-\alpha$ —that this upper bound will not underestimate the true transition point. In other words, choosing the upper bound ensures a conservative decision that maintains the user-specified confidence level.*

### 4.5 ERGo Overhead

ERGo implements a transition mechanism from a full-parameter to a parameter-efficient phase in training GNNs on PIM accelerators. The transition point (*t*) is predetermined during the pre-processing phase using the conformal calibrated regression model $C$ (line 4 of Algorithm 2). Algorithm 2 details the end-to-end hybrid fine-tuning procedure employing ERGo for training a GNN model ($gnn_i$) with graph data (*G*) on a PIM hardware accelerator ($Acr$). Initially, the full-parameter phase is utilized till the training epochs reach *t*. Afterwards, the mapping for the parameter-efficient phase is utilized for the remainder of the training process. This transition necessitates an update to the computational kernel-to-core mapping on the hardware. The updated mapping is facilitated through on-chip registers within the control logic of each PE (see Fig. 2). First, the trainable GNN weights are frozen and additional LoRA parameters are initialized. In homogeneous accelerators, this results in all matrices being mapped to the same type of compute resources (line 20 of Algorithm 2). As discussed previously, the advantage of reducing off-chip access in ERGo through LoRA is equally applicable to both homogeneous and heterogeneous accelerators. The additional benefit of ERGo in a heterogeneous NVM-SRAM accelerator is the improvement in lifetime of NVM cells by preventing rewrites. In architectures where the accelerator consists of both NVM ($Acr_{NVM}$) and SRAM ($Acr_{SRAM}$) PEs, we utilize the mapping discussed in subsection 4.3 where trainable LoRA parameters are mapped to SRAM PEs. For the NVM PEs in the accelerator ($Acr_{NVM}$), this results in freezing of the trainable GNN weights. Conversely, in the SRAM PEs ($Acr_{SRAM}$), it introduces additional computation in the forward pass, specifically MVM operations involving input and LoRA parameters (line 17 of Algorithm 2). However, this increase in forward pass computation is offset by a reduction in backward pass computation on SRAM, as discussed in subsection 4.3. Specifically, gradients are calculated solely for the LoRA parameters, which significantly reduces the number of MVM operations, given $r \ll d$ (Algorithm 2, Line 25). Overall, the ERGo framework does not introduce additional overhead

ALGORITHM 2: ERGo

**Input**: $gnn_i$ = GNN model to be trained, $G$ = Graph dataset, $n$ = Training epochs, $C$ = CP model, Acr = PIM Accelerator

**Output**: $gnn_i$ = Trained GNN model

```
/* Pre − processing (host) */
∑_{j=1}^{p} (SG_j) ← G                  /* Partition the graph G into p subgraphs */
X_i ← Features{gnn_i, G}                /* Extract the feature set of the GNN instance */
[lb, ub] ← C(X_i)                       /* Determine upper and lower bound for transition point using CP */
transition point (t) ← ub               /* Set ub as transition point to ensure (1 − α)% coverage */
/* Training (Accelerator) */
Initialize gnn_i
while epoch < n do
    foreach SG_j ∈ G do
        if epoch < t then
            /* Full-parameter phase */
            if NVM PEs ∈ Acr and CMOS PEs ∈ Acr then
                /* Mapping on heterogeneous Acr */
                Map(Acr_NVM) ← W ∈ {W^1, W^2, …, W^l}             // GNN weights to NVM PEs
                Map(Acr_SRAM) ← Acr_SGj                            // Adjacency matrix to SRAM PEs
            else
                /* Mapping on homogeneous Acr */
                Map(Acr) ← W ∈ {W^1, W^2, …, W^l}, Acr_SGj
        else
            /* Parameter-efficient phase */
            W_O ← Freeze(W ∈ {W^1, W^2, …, W^l})                   // Freeze GNN weights
            if NVM PEs ∈ Acr and CMOS PEs ∈ Acr then
                /* Mapping on heterogeneous Acr */
                Map(Acr_SRAM) ← A, B ∈ { {A^1, A^2, …, A^l}
                                         {B^1, B^2, …, B^l} }      // LoRA weights to SRAM PEs
                Map(Acr_SRAM) ← Acr_SGj
            else
                /* Mapping on homogeneous Acr */
                Map(Acr) ← A, B, Acr_SGj
        Train(gnn_i, SG_j, Acr)                                    // Train the GNN model on the accelerator
        if epoch < t then
            Acr ← Weight_update(W)                                 // Update weights
        else
            Acr ← Weight_update(A, B)                              // Update LoRA weights
    end for
end while
return gnn_i                            /* Trained GNN model */
```

during the transition phase. Instead, it primarily leads to a reconfiguration of the compute mapping on the hardware platform to enhance both performance and energy efficiency.

Importantly, ACE leverages key advantages inherent to PIM architectures. The frozen parameters remain on-chip in PIM PEs throughout the fine-tuning process. This contrasts with the LoRA implementation on computing platforms such as GPUs. In GPU-based systems, both frozen and LoRA parameters are fetched on-chip for computation during fine-tuning. The primary reason for employing LoRA on GPUs is to enable larger models or batch sizes to fit within existing memory constraints [8] [9]. It may introduce timing overhead due to an increase in the total parameter count (frozen + trainable) and the additional computation with the LoRA parameters.

## 5 EXPERIMENTAL RESULTS

In this section, we present an experimental analysis of ERGo for GNN training on PIM architectures. First, we describe the experimental setup used for the performance evaluation. Next, we assess the effectiveness of ERGo with conventional full-parameter training in terms of predictive accuracy. Finally, we perform endurance, performance, and energy analysis of ERGo in comparison to the baselines.

### 5.1 Experimental Setup

**Graph datasets -** We consider five graph datasets in our evaluation: protein-protein interaction (PPI), Reddit, Open Graph Benchmark-citation2 (Ogbl), Cora, and Amazon [39]. These graph datasets encompass a broad spectrum of graph sizes, representing the typical size and diversity of use cases for training GNN models at the edge.

**GNN Models -** We also utilize three types of GNN models: Graph Convolution Networks (GCN), Graph Attention Networks (GAT), and Graph Sample and Aggregate (SAGE) [40] [41] [42]. Moreover, to demonstrate the generalizability of ERGo, each graph dataset is paired with a respective GNN model in our experimental evaluation, though any model can be applied to any graph dataset. Table 1 presents the details of the datasets, and hyperparameters such as learning rate (lr), batch size (Batch), number of graph partitions (p), and training epochs (e).

**Configuration** - We employ mini-batch training, which uses subgraphs obtained from a monolithic graph via partitioning. Here, the graph is partitioned into smaller subgraphs using the METIS tool [43]. The partitioning step is executed once per graph dataset during the preprocessing phase on the host device and takes a small portion of time (less than 1%) for the datasets considered here. As discussed in Section IV, we utilize the calibrated trained regression model to determine the transition point in ERGo. We first generate the necessary training/calibration data (see *creation of augmented GNN training instances part* in the previous section), utilizing Reddit and PPI datasets to train and calibrate the regression model. We vary the number of node features and the number of subgraphs to create one hundred diverse augmented GNN training instances and find the true transition point for each training task. Note that the complete set of node features and subgraphs from both the Reddit and PPI datasets is not used for the training data, as these serve as part of the test data. We employed 80% of this data for training the regression model and a different 20% data for calibrating the model using CP for regression with $\alpha$=0.05 (i.e., 95% coverage) parameter. Both the training of the regression model and the calibration process are one-time procedures and are not part of the execution flow under deployment. Subsequently, the calibrated model is employed to determine the transition point for an unseen GNN dataset on the host device during the pre-processing phase, after graph partitioning. The prediction using the calibrated regression model adds a negligible performance overhead (~0.01% ) in the end-to-end GNN training process.

We apply LoRA to GNN weights during the parameter-efficient phase. Consistent with prior work [9], we conduct a sweep of the rank (r) parameter, varying its values from 8 to 128 [9]. We employ a modified NeuroSim v2.1 to obtain the area, latency, energy of PIM tiles, all on-chip buffers, and peripheral circuits. Additionally, we take into account the overhead associated with loading weights and data access to/from the DRAM for reporting performance and energy metrics of each architecture. An industry-standard High Bandwidth Memory (HBM2) is used as the DRAM for off-chip data access [44]. A 16-bit fixed-point precision was utilized for the storage and computation of the model parameters and activations during training. The 16-bit fixed-point arithmetic provides adequate numerical precision for effective GNN training from scratch.

Table 1: Graph datasets and GNN workload configuration

| Dataset | Dataset Statistics (Nodes, Edges) | Hyperparameters | GNN Model |
|---|---|---|---|
| CoRA | (2708, 5429) | Batch=1, p=5, e=200, lr =0.01 | GAT |
| PPI | (56.94 K, 818.71 K) | Batch=5, p=250, e=100, lr =0.01 | GCN |
| Reddit | (232.96 K, 11.606 M) | Batch=10, p=1500, e=50, lr =0.01 | SAGE |
| Amazon | (2.449 M, 61.859 M) | Batch=20, p=10000, e=100, lr =0.01 | GCN |
| Ogbl | (2.927 M,30.561 M) | Batch=16, p=15000, e=50, lr =0.01 | SAGE |

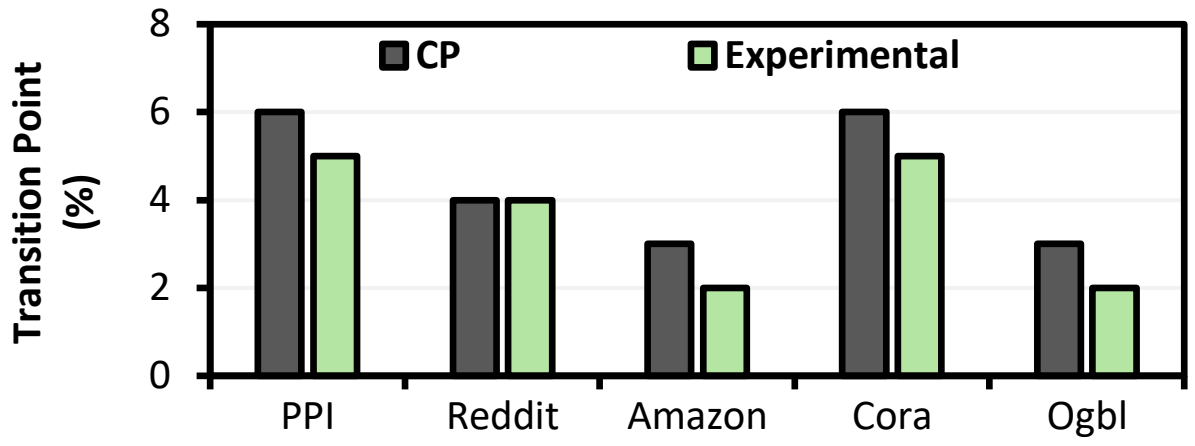

Figure 5: The transition point determined in ERGo using the calibrated regression model and the true transition point determined experimentally for different GNN datasets.

### 5.2 ERGo Accuracy Analysis

In this section, we assess the accuracy of ERGo and compare it with full-parameter training (FpT). We first must determine the appropriate transition point in ERGo. Fig. 5 presents the transition point determined via conformal prediction (CP) for each of the GNN datasets presented in Table 1. Additionally, we experimentally determine the true transition point to achieve less than 1% accuracy loss (ground truth). For example, in the case of the Amazon dataset, the CP-enabled prediction for the transition point in Fig. 5 occurs at 3% of the total training epochs, whereas the experimentally derived result (Experimental) indicates that the true transition point is at 2%. From Fig. 5, it is evident that CP correctly bounds the transition point required to preserve accuracy in ERGo for each GNN dataset considered with the difference between predicted transition point and true transition point less than 1% of training epochs.

Next, we investigate the impact of different LoRA ranks (r) on GNN model accuracy. Figs. 6(a) and 6(b) illustrate the model accuracy with different ranks for Reddit and Amazon datasets, respectively. Here, "LoRA-n" denotes the rank utilized for representing the GNN weights during the parameter-efficient phase of training. Note that a lower rank is preferred since it reduces the total trainable parameters. From the results presented in Fig. 6, we see that a rank of 8 $(\mathrm{LoRA}-8)$ results in a significant drop in accuracy, up to 6%, making it not suitable for the parameter-efficient phase. In contrast, ranks of $r = 32$ and $r = 64$ exhibit minimal drop in accuracy. Overall, we observe that model accuracy increases as we increase the rank of LoRA matrices. Notably, there is no observed improvement in accuracy beyond $r = 32$, as the accuracy remains unchanged at r = 64. Hence, we do not utilize ranks beyond 32 for the models considered in our evaluation since it leads to an increase in trainable parameters with no gain in accuracy. In summary, based on our ablation analysis, a rank of 16 can be utilized when minimizing training cost is the primary objective and an accuracy loss of up to 2% is tolerable. Conversely, if maintaining accuracy is of primary concern, a rank of 32 is more appropriate. In this work, we adopt r = 32 for all the results presented later in the experimental section.

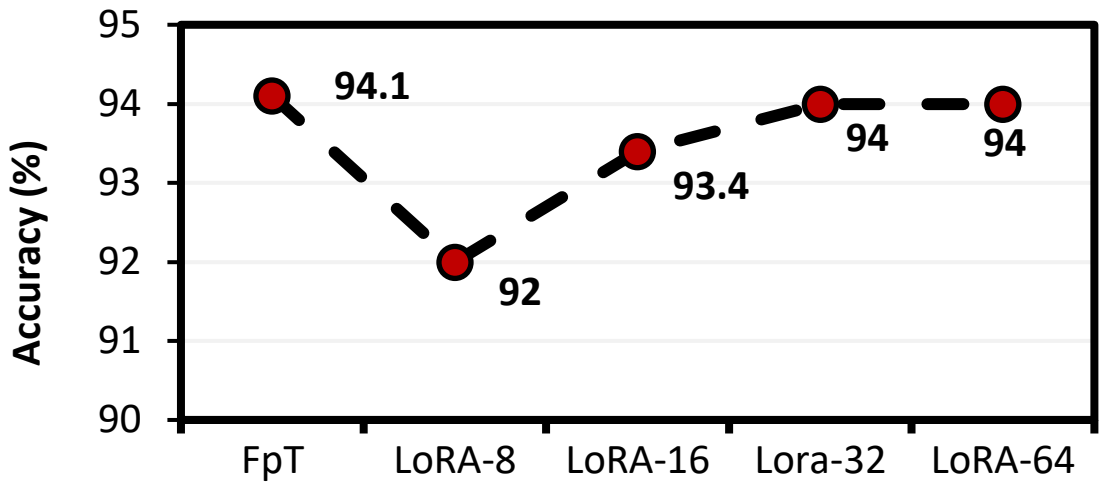

(a) Reddit

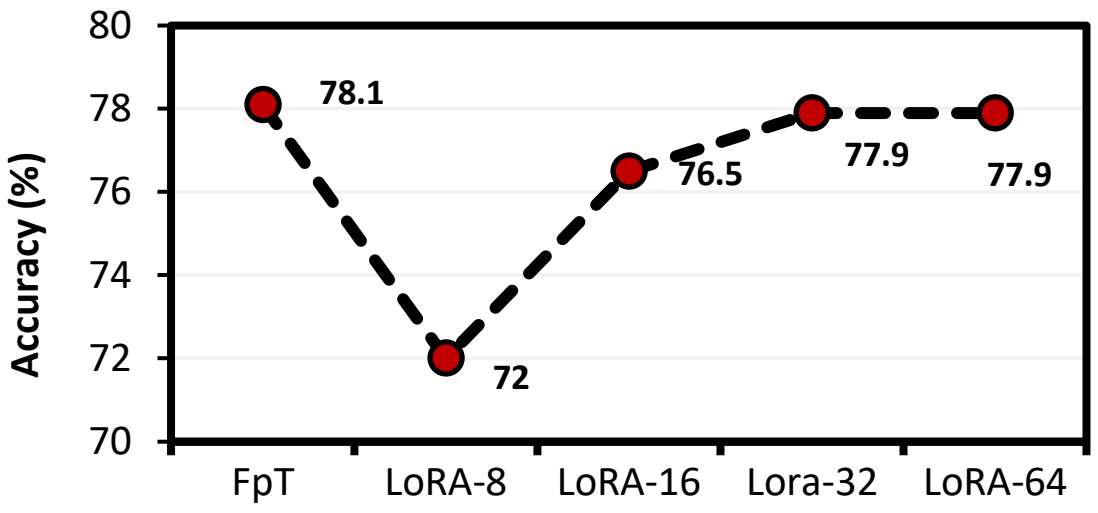

(b) Amazon

Figure 6: Comparative Analysis of GNN model accuracy under different LoRA ranks for reddit and amazon datasets as examples.

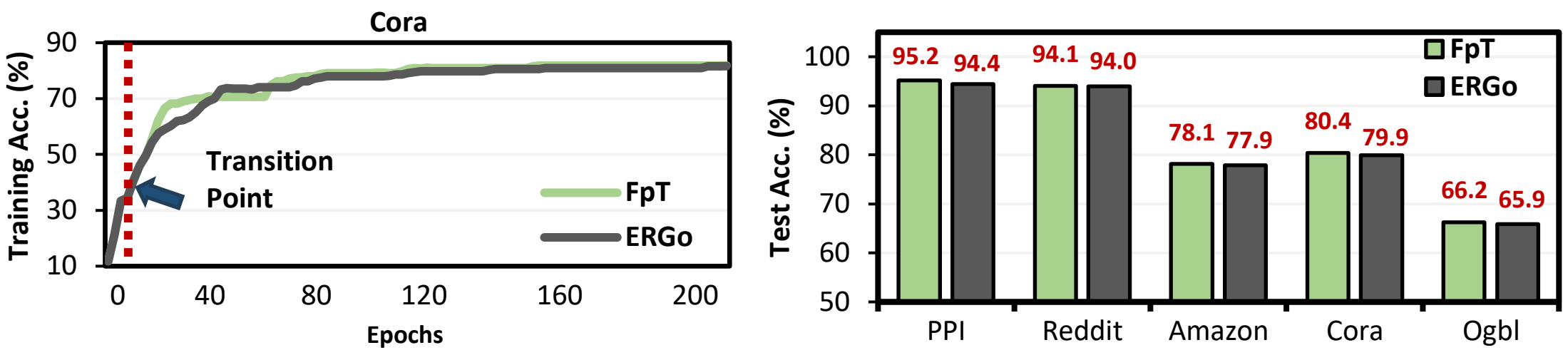

(a) (b)

Figure 7: Comparative analysis of ERGo in comparison to FpT showing, (a) training accuracy plot across epochs and (b) test accuracy.

Now, we evaluate the accuracy of ERGo compared to FpT across all datasets. Fig. 7 (a) illustrates the training accuracy plot of Cora as an example. The transition point in ERGo for Cora occurs at 6% (as shown in Fig. 5), which corresponds to epoch 12. From Fig. 7 (a), we observe that ERGo achieves training accuracy comparable to FpT, demonstrating the effectiveness of a hybrid framework in training GNNs from scratch. Now, Fig. 7 (b) presents the test accuracy for ERGo and FpT respectively, across the diverse set of GNN datasets outlined in Table 1. Note that we keep the total training epochs consistent across ERGo and FpT in our analysis. In Fig. 7 (b), we observe an average accuracy loss of 0.38% with the worst case accuracy loss of 0.8% for PPI (within the user-specified margin) in comparison to FpT. These findings suggest that ERGo can train GNNs with little to no loss in accuracy when compared to FpT. It should be noted that the drop in accuracy for datasets is between 0.1% and 0.5%, except for PPI, where the accuracy loss is 0.8%. Even for PPI, we were able to reduce the accuracy loss to 0.2% by training the regression model using an output label corresponding to a 0.5% accuracy loss, instead of 1% loss. This results in a delayed transition to the parameter-efficient phase, shifting from the 6th epoch to the 11th epoch, which will lead to a performance penalty.

Finally, we examine the effectiveness of ERGo within a typical ablation scenario during GNN training. The CP training and calibration process outlined in Section 4.4 leverages generic features of the GNN training instance to determine the transition point. Specifically, we utilize parameters such as the number of features per node, the number of subgraphs, the number of edges, and the number of output labels targeted by the prediction task. Importantly, these features remain consistent even as hyperparameters or training strategies (e.g., learning rate, optimizer, or batch size) are varied during ablation for a training task. For this purpose, we consider an experiment where we vary hyperparameters -including learning rate (0.001 and 0.0005), optimizer (Adam and SGD), and batch size (20 and 30)- as an example [45] [46]. Table 2 presents the accuracy achieved by both the baseline FpT and ERGo for each configuration over 50 epochs on the Reddit dataset as a representative example. We observe that ERGo consistently maintains accuracy within the 1% allowable margin of FpT across all hyperparameter configurations examined. Note that the training of the regression model uses an output label of 1% accuracy loss here. The results demonstrate the generalizability of our method, showing ERGo remains effective under common ablation scenarios.

Table 2: Model Accuracy with FpT and ERGo using various hyperparameters for the Reddit dataset

| | Batch=20 | | Batch=30 | |
|---|---|---|---|---|
| Optimizer, Learning Rate | FpT | ERGo | FpT | ERGo |
| Opt=Adam, Lr=0.01 | 93.59% | 93.40% | 93.27% | 93.40% |
| Opt=Adam, Lr=0.005 | 94.08% | 93.92% | 94.10% | 93.80% |
| Opt=SGD, Lr=0.01 | 94.34% | 93.90% | 94.10% | 94.00% |
| Opt=SGD, Lr=0.005 | 93.70% | 93.50% | 93.80% | 93.40% |

### 5.3 Intermediate Activations utilizing ERGo

A significant amount of intermediate activations is generated during the training process. For instance, using the Reddit dataset as an example, the forward activations generated for a single layer are approximately 50 MB in size when considering 16-bit fixed-point precision. To enable backpropagation, activations from all forward layers (three layers in the case of Reddit) must be stored alongside errors and gradients for the complete batch. This would require over 500 MB of on-chip storage to process each batch. It would result in significant area and power overhead to provide sufficient on-chip memory to store all activations during the training process. Existing works have highlighted the need to consider off-chip storage while utilizing PIM architectures for training, including homogeneous, heterogeneous, digital, and analog implementations [23] [47]. However, the cost of off-chip DRAM access is orders of magnitude higher than that of computation in terms of latency and energy.

In this section, we compare ERGo with the baseline, FpT, focusing on intermediate activation size. The mini-batch training performs the following computations for processing each batch: forward pass, error calculation, gradient computation, and weight update. Fig. 8 presents the normalized intermediate activations for each computational kernel during the training process, compared to the baseline FpT. We use the Reddit dataset as a representative example for this analysis. During mini-batch training, the forward pass, error propagation, and gradient computation occur for each input subgraph in the batch, while the weight update is performed only once per batch. Consequently, we report the normalized size of activations for processing one complete batch, referred to as "Total" in Fig. 8. We observe that ERGo does not result in any reduction of activations for the forward pass. This is because the forward activations generated for each layer, which is the sum of the activations from the frozen and the LoRA branch have the same output dimension. Similarly, the dimensionality of errors generated for each layer remains unchanged. However, ERGo significantly reduces the amount of intermediate activations during back-propagation computation. The implementation of LoRA during the parameter-efficient phase decreases the number of trainable parameters and associated gradients. Furthermore, weight updates are limited to the LoRA matrix, which is substantially smaller than the frozen matrix. It is evident from the results presented in Fig.7 and Fig. 8 that ERGo reduces intermediate activation size without sacrificing accuracy.

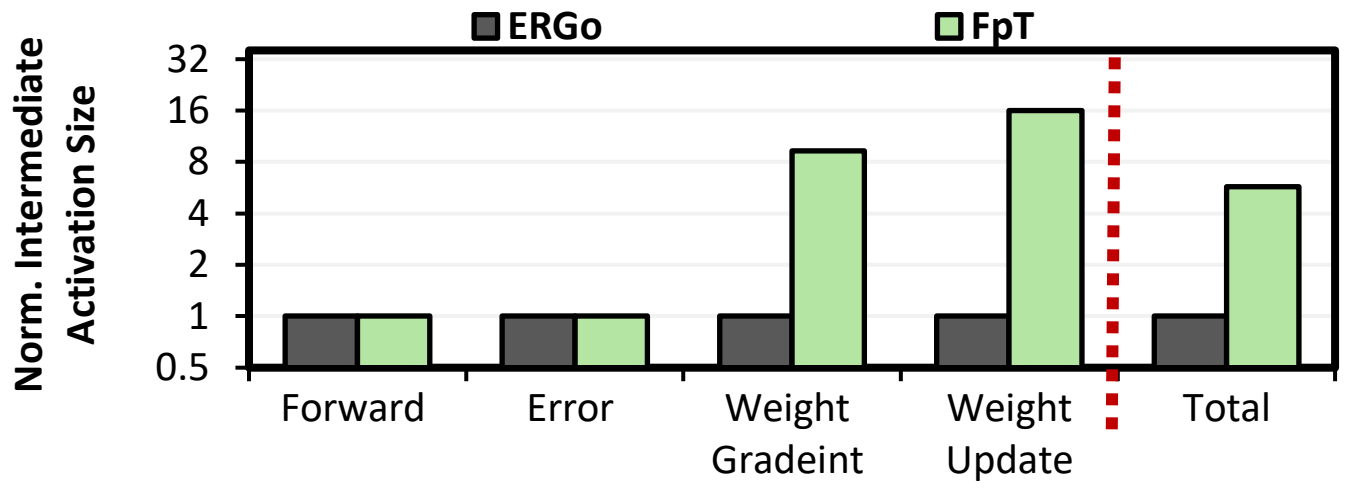

Figure 8: Normalized intermediate activation size for the baseline FpT against ERGo, considering each computational kernel during GNN training on the Reddit dataset.

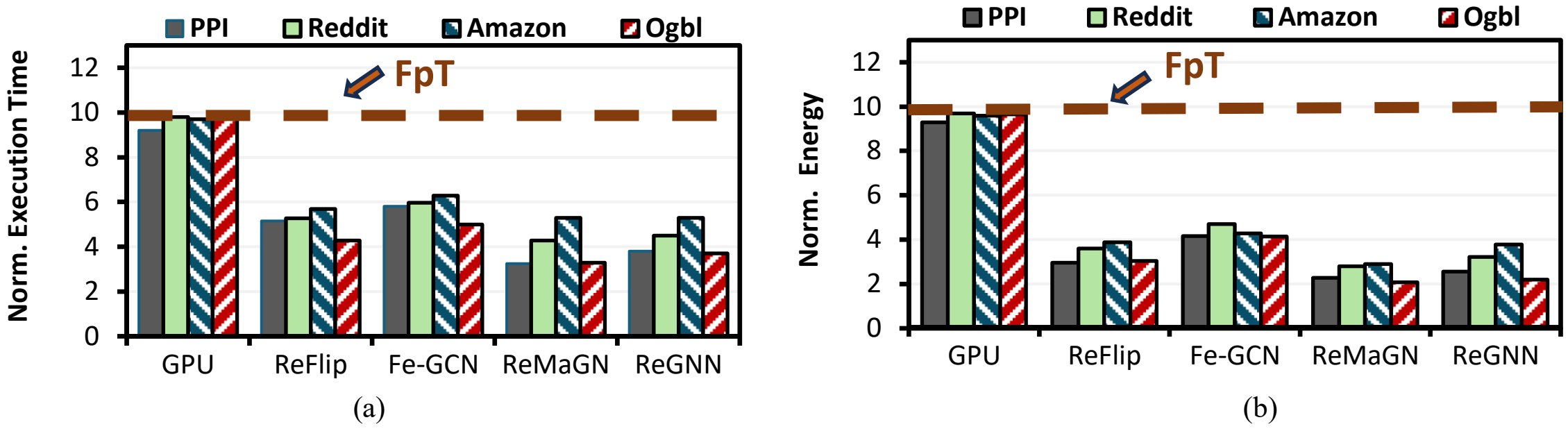

Figure 9: Normalized (a) execution time and (b) energy respectively of ERGo when compared to FpT on the same hardware platform across various GNN datasets.

### 5.4 ERGo Performance Evaluation

The advantage of ERGo in reducing the off-chip memory accesses during training is independent of the underlying PIM platform. We now evaluate the performance of ERGo using a few representative PIM-based GNN accelerators. We consider ReFlip, Fe-GCN, ReMaGN, and ReGNN as examples for this analysis [4] [16] [17] [19]. ReFlip is a ReRAM-based accelerator featuring 64×64 crossbar configurations. It comprises 64 PEs, each containing 8 tiles with 32 crossbars arranged in a 4×8 matrix layout per tile [19]. Similarly, ReMAGN is also a ReRAM-based accelerator for GNN training. ReMaGN employs homogeneous crossbars of dimension 128×128 for both combination and aggregation kernels [4]. In contrast, ReGNN employs a combination of ReRAM-based digital and analog PIM for GNN acceleration. The analog PIM utilizes a 128×128 crossbar configuration, while the digital PIM employs 1024×1024 configuration [17]. All these PIM architectures share a similar on-chip memory hierarchy, as depicted in Fig. 2, but differ in their on-chip memory capacities. To ensure consistency, we maintain the same DRAM data access and energy parameters across our evaluation. Additionally, we include Fe-GCN, which is a homogeneous NVM accelerator but utilizes FeFET as the NVM device for GNN acceleration. Unlike others, Fe-GCN integrates ternary content-addressable memory (TCAM) to store data along with FeFET crossbars with dimension of 128×128 for computation [16]. We also include Nvidia Tesla V100 GPU as a baseline in this analysis.

We adopt the hybrid training dataflow on each accelerator, without introducing additional hardware resources. Figs. 9(a) and 9(b) present the normalized execution time and normalized energy, respectively for ERGo compared to FpT on the same hardware platform. From Fig. 9(a) and Fig. 9(b), we observe that hybrid training does not lead to any noticeable speedup or improved energy efficiency for GPU systems. This limitation arises because GPUs store frozen matrices off-chip, requiring frequent retrieval of both frozen weights and LoRA parameters from off-chip memory during computation in the parameter-efficient phase. Importantly, this is not the case for PIM platforms. PIM platforms store frozen parameters on-chip within PIM crossbars, necessitating only off-chip storage for LoRA parameters. This presents the distinct advantage of employing LoRA when applied to PIM platforms. Furthermore, hybrid training reduces computation associated with gradient calculation during backpropagation computation, a key bottleneck in PIM devices, resulting in performance gains. Among the PIM architectures evaluated, we see significant performance improvement with ERGo compared to FpT, highlighting the advantage of ERGo to existing architectures. The gain in energy efficiency is higher compared to performance for ERGo, as unlike off-chip latency, the energy cost of data access cannot be hidden or reduced through pipelining. For instance, the activations can be prefetched from off-chip buffers since PIM executes computation in-order with deterministic latencies. We observe that ReMAGN achieves the highest speedup, followed by ReGNN,

ReFlip, and Fe-GCN, respectively. These performance and energy-efficiency gains are closely tied to the on-chip memory capacity of each accelerator. For instance, ReFlip features a 256 KB buffer per PE, resulting in a total of 16 MB of on-chip memory across its 64 PEs. In comparison, ReMAGN has only 8 MB of on-chip buffers. Consequently, ReMAGN achieves higher energy-efficiency gains with ERGo due to its smaller memory footprint. Fe-GCN achieves the least gain since it utilizes TCAM for storing data. The energy to fetch data from TCAM is lower in comparison to off-chip DRAM. Furthermore, it is critical to emphasize that none of the PIM platforms analyzed possess sufficient on-chip buffers (cache) to store all the intermediate activations. As discussed in the previous section, we would need over 500 MB of on-chip buffer storage to eliminate memory access. However, providing such large on-chip buffers is infeasible due to the area and power-related overheads. ERGo is designed to address these constraints without compromising the generalizability of the approach by reducing both the trainable parameters and off-chip memory access.

### 5.5 ERGo Lifetime Analysis

The number of write operations to NVM crossbars directly influences the lifetime of NVM cells. In ERGo, the parameter-efficient phase freezes the weights from the full-parameter phase and uses LoRA matrices as trainable parameters. As discussed in Section 4, employing SRAM instead of NVM for mapping LoRA matrices can prevent rewrites due to weight updates on NVM crossbars. This mapping can consequently enhance the lifetime of NVM cells significantly, since ERGo utilizes a parameter-efficient phase for most of the training (greater than 94%). For this purpose, we consider a heterogeneous NVM-SRAM PIM architecture as the hardware platform and evaluate the benefit in lifetime improvement by utilizing ERGo for GNN training. However, manufacturing technologies of NVM crossbars may not be compatible with that of SRAM cells. Three-dimensional (3D) integration is a practical solution for heterogeneous architecture since it can support several types of devices fabricated using different technologies in a single system. In this work, we use a 3D system consisting of four planar tiers as illustrated in Fig. 10, each of size $10\text{ mm} \times 10\text{ mm}$, noting that any other system configuration can be used without affecting the generalizability of our findings.

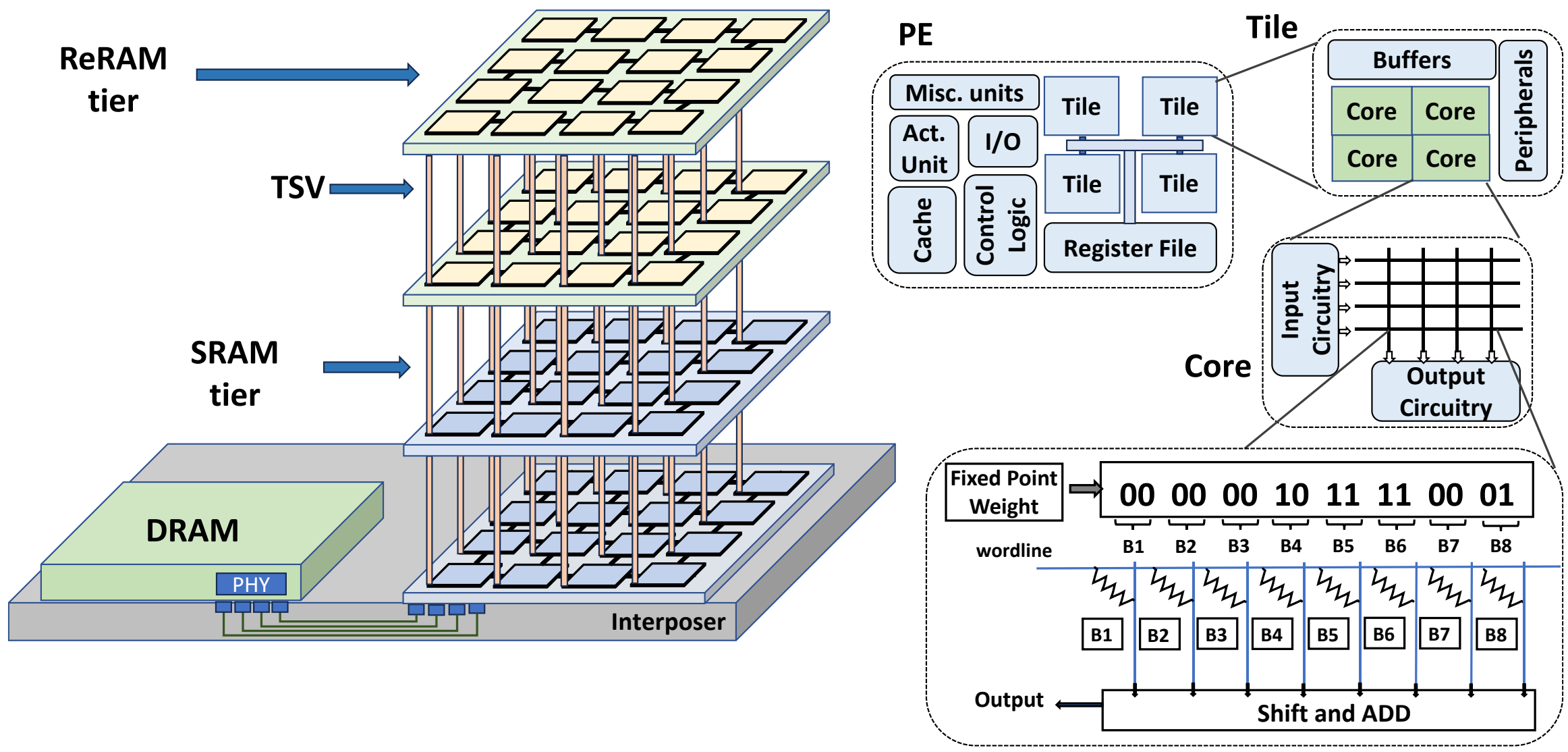

Figure 10: ReRAM-SRAM heterogeneous architecture consisting of 2 ReRAM and 2 SRAM tiers respectively connected vertically through TSV-based interconnect. The DRAM is connected through the interposer via 2.5D integration. The die heights are not drawn to scale. This figure is for illustration purposes only

Table 3: PIM architecture specifications

| 64 PEs (16 PEs/tier), 4 tiles/PE | |
|---|---|
| ReRAM Tile (Analog PIM) | 96 ADCs (8-bits), 128×96 DACs (1-bit), 96 crossbars, 128×128 crossbar array, 2-bit/cell resolution, IT1R cell, 64 KB eDRAM, 0.40 $mm^2$ |
| SRAM Tile (Digital PIM) | 256 S/A, 8KB SRAM array (256×256), 9 column/row-de coder, 9 SRAM arrays, 1-bit-cell, 8T cell, 64 KB eDRAM, 0.40 $mm^2$ |

Table 4: Design parameters

| | |
|---|---|
| TSV Parameters | Diameter – 5 µm, Pitch – 15 µm, Via Height – 15 µm, Capacitance – 37 fF, Resistance- 20 mΩ [50] |
| DRAM Parameters | HBM2, 2 GB per tier, 256 GB/sec max bandwidth, 1024-bit memory interface, 600 MHz clock [44] |

A pipelined approach is utilized in the forward pass to optimize throughput, consistent with prior work [23]. In contrast, the backward pass is performed sequentially, one layer at a time, given the memory footprint required to process each layer [25]. The throughput of the pipelined system in the forward pass is constrained by its slowest component. The forward pass utilizes both NVM and SRAM PEs. Hence, we balance the stage delay in the pipelined implementation and allocate resources appropriately among ReRAM and SRAM PEs [47]. Specifically, we consider two tiers for ReRAM PEs and one tier of SRAM PEs for the forward-pass computation. Note that both the SRAM and ReRAM PEs cannot be integrated on the same tier due to technology limitations. Additionally, one SRAM tier is dedicated to support back-propagation computation. Overall, we have considered a system with 2 NVM and 2 SRAM tiers. Fig. 10 illustrates a conceptual NVM-SRAM PIM architecture consisting of vertically integrated tiers, each with either NVM or SRAM PIM. The planar tiers are connected vertically with Through Silicon Via (TSV)-based links using hybrid bonding [48] [49] [50]. We employ ReRAM as the NVM device; however, other NVM technologies can also be used with ERGo without altering the observations. Table 3 provides the specifications of the ReRAM and SRAM PIM tiles. A multi-level cell is used to represent the parameters stored on ReRAM tiles. Specifically, the ReRAM cells utilize a 2-bit per cell resolution, resulting in eight cells to represent 16-bit fixed-point weights. Fig.10 (bottom right) illustrates the 2-bit per cell representation utilized on ReRAM crossbars. In contrast, SRAM employs a 1-bit per cell representation necessitating 16 cells to represent each weight value. Subsequently, partial outputs are accumulated using a shift-and-add operation to obtain the final output of the MVM operation. The total on-chip memory to store the activations is 4 MB. GNN training utilizes the same adjacency matrix to perform an aggregation operation across all the layers. This results in many-to-few and few-to-many traffic patterns between PEs. Hence, we utilize a multicast-enabled 3D mesh topology as the interconnection backbone to connect PEs and support the traffic generated during GNN training [4]. Table 4 shows the DRAM parameters and physical parameters for the TSV-based vertical links. The DRAM access to store the intermediate activations is facilitated through an interposer via 2.5D integration [51].

We now quantify the worst-case number of write operations to NVM cells when using ERGo, comparing it to the FpT approach on the heterogeneous architecture presented in Fig. 10. ERGo does not lead to any lifetime improvements for homogeneous NVM-based accelerators since LoRA matrices are also mapped onto NVM, which gets updated during training. We consider an endurance of $10^8$ write cycles for ReRAM cells following existing literature [26]. Note that SRAM does not have this limitation of low device endurance. We also include a comparison with homogeneous NVM-based architecture discussed in the previous subsection (ReFlip, Fe-GCN, ReGNN, and ReMAGN), collectively referred to as NVM-only architectures. Additionally, we examine ERGo against lifetime improvement techniques designed to reduce the number of writes during DNN training. Specifically, we employ Structured gradient sparsification (SGS) as one of our baselines for this analysis [35]. As discussed in Section 2, SGS has been shown to achieve lifetime improvements by performing gradient accumulation and pruning gradients based on their magnitude, thereby reducing the frequency of weight updates. Moreover, we consider an enhancement of SGS, which combines SGS with row-swapping (SGS+RS) to distribute writes evenly across the crossbar, serving as an additional baseline [35]. Both SGS and SGS+RS reduce the number of write operations only due to weight updates and do not address the writes due to activations and the adjacency matrix. Thus, we integrate two SGS variants with FpT (FpT-SGS and FpT-SGS+RS) on the heterogeneous architecture, which leverages SRAM for aggregation and back-propagation. Note that the worst-case number of writes remains the same when SGS is incorporated in NVM-only architectures since the NVM crossbars will have writes due to the adjacency matrix and activations.

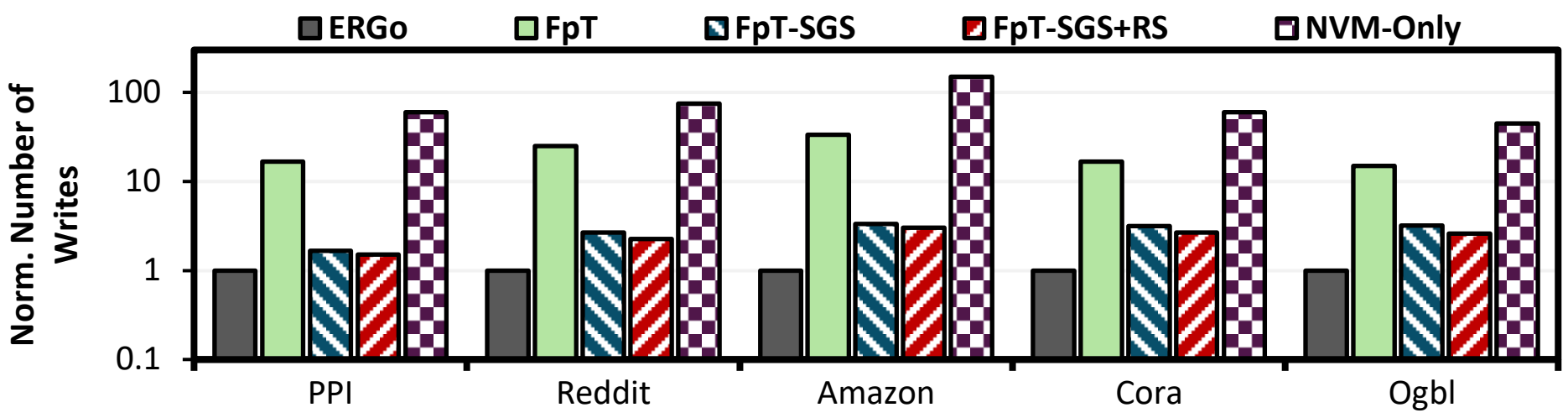

Figure 11: Normalized number of write operations to NVM cells during training when compared to ERGo for the baselines across different GNN datasets.

Fig. 11 illustrates the normalized worst-case number of writes using the above-mentioned techniques with respect to ERGo across GNN datasets detailed in Table 1. We observe that ERGo results in a much lower number of writes in comparison to the baselines. This can be attributed to the freezing of weights on NVM during the parameter-efficient phase, which constitutes most of the training in terms of epochs (over 94% training epochs) for the GNN datasets evaluated. For instance, Reddit dataset with FpT necessitates $7.5\times10^3$ write operations given the hyperparameters presented in Table 1 of the manuscript. In contrast, ERGo necessitates only $3\times10^2$ write operations leading to a 25× reduction in the total number of writes due to weight updates during the end-to-end training process. The reduction in writes is most pronounced in the Amazon dataset (33 ×) for ERGo, as it transitions to the parameter-efficient phase at the earliest. FpT-SGS also helps reduce the worst-case number of writes on the heterogeneous architecture. However, it leads to performance degradation as we show later. The combination of row swapping with SGS (FpT-SGS+RS) yields only marginal improvements over FpT-SGS in reducing the number of writes. In contrast, NVM-only results in the highest number of writes since back-propagation computation is performed on NVM crossbars, which necessitates multiple write operations in each batch. The results in Fig. 11 show the normalized number of write operations during training. However, Fe-GCN, which is a FeFET-based accelerator, will fail earlier in comparison to ReRAM-based homogeneous counterpart, since typical FeFET device endurance ($10^5$) is less in comparison to the ReRAM cell endurance ($10^8$) [26].

We now compare ERGo to the baselines, FpT, FpT-SGS, and FpT-SGS+RS in terms of off-chip DRAM data access and Energy-Delay Product (EDP) on the heterogeneous ReRAM-SRAM hardware platform considered in Fig. 10. We use EDP as the evaluation metric since it captures both performance and energy in a single term. Fig. 12 (a) shows the normalized amount of data accessed through DRAM for the Reddit dataset as an example when utilizing ERGo. From Fig.

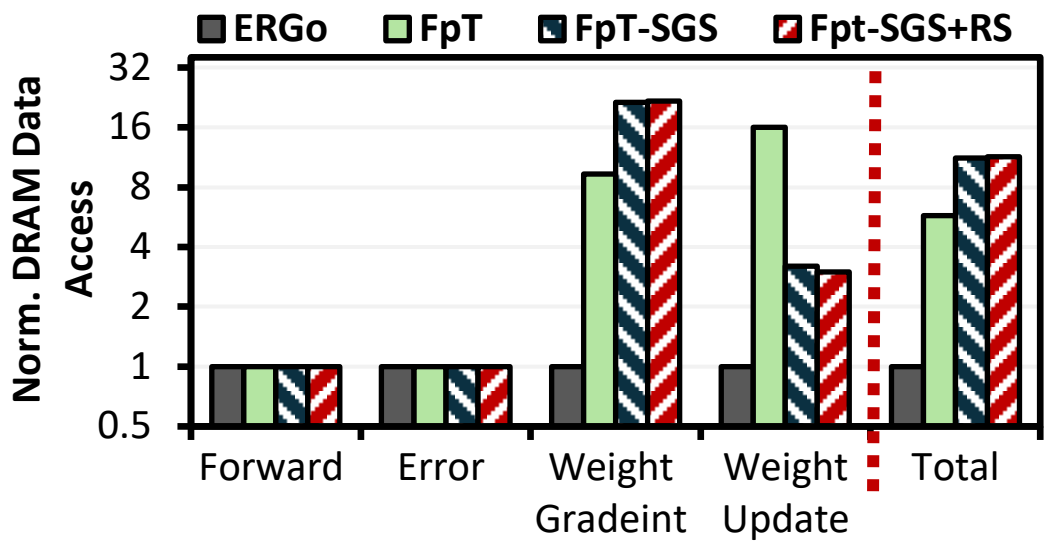

(a)

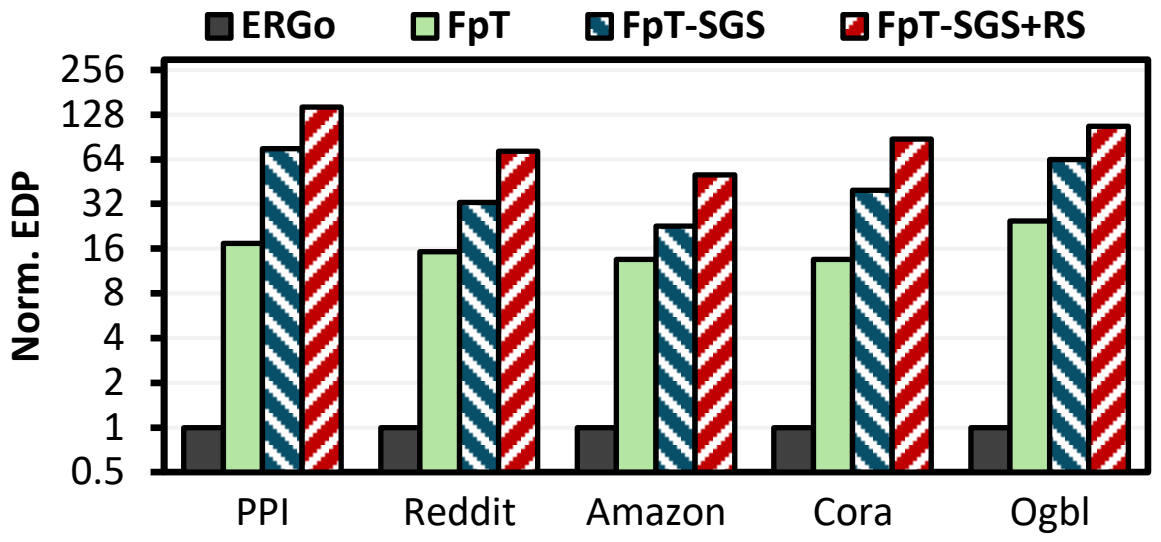

(b)

Figure 12: Normalized (a) DRAM data access for the Reddit dataset and (b) Energy-Delay Product (EDP) across various GNN datasets, respectively, for the baselines on the heterogeneous hardware platform presented in Fig. 10.

12 (a), we observe that ERGo significantly reduces the amount of off-chip access during the back-propagation computation. In contrast, FpT-SGS and FpT-SGS+RS exhibit a substantial increase in the size of intermediate activations during gradient computation. This is due to the accumulation of gradients, leading to higher intermediate activation storage requirements. Finally, Fig. 12 (b) presents the EDP for the baselines with respect to ERGo for the GNN datasets outlined in Table 1. A similar trend is observed in the EDP metric, where ERGo outperforms the baselines, as shown in Fig. 12 (b). The improvements for ERGo correlate with the reduction in off-chip access. Notably, both FpT-SGS and FpT-SGS+RS improve the lifetime of NVM cells (as shown in Fig. 11) at a significantly higher cost in terms of EDP when compared to ERGo. Overall, ERGo can help in improving the lifetime of NVM devices by an order of magnitude and lead to superior performance and energy-efficiency without introducing overheads.

### 5.6 ERGo EDP vs Accuracy Tradeoff

ERGo through the CP-based approach determines the transition point from the full-parameter phase to the parameter-efficient phase for each GNN training task. Specifically, the calibrated regression model identifies the switching epoch based on intrinsic characteristics of the dataset (e.g., graph size, node features, and complexity of the task), enabling it to adapt to different GNN architectures and datasets automatically. The epoch at which this transition occurs directly influences the overall efficiency of the training process and the model accuracy.

We now compare our CP-driven approach against arbitrary fixed-threshold approaches for determining the transition point in ERGo. Table 5 presents the EDP gain achieved by ERGo compared to FpT with CP-based and arbitrarily determined transition points (10% and 15% of training epochs) for the heterogeneous architecture illustrated in Fig. 10. Note that the transition point presented in Figure 5 for each dataset is used for reporting EDP values with the CP-based approach. From Table 5, we see a significant advantage of utilizing the CP-based approach over the fixed selection of transition points. The difference is particularly notable for the Ogbl dataset, where the CP-driven method achieves approximately 2.90× higher EDP gain compared to an arbitrary threshold of 15%. This shows that while arbitrary methods might occasionally yield reasonable results, they lack predictability and generalizability, often requiring empirical tuning for each scenario.

Table 5: Comparative Analysis of CP-based vs. Arbitrarily Determined transition points in ERGo

| | **Transition Point (CP)** | **Transition Point – 10%** | **Transition Point – 15%** |
|---|---|---|---|
| | EDP Gain | EDP Gain | EDP Gain |
| PPI | 17.31 × | 11.95 × | 8.61 × |
| Reddit | 15.29 × | 9.71 × | 7.44 × |
| Amazon | 13.53 × | 8.57 × | 6.79 × |
| Cora | 13.52 × | 10.08 × | 7.65 × |
| Ogbl | 24.63 × | 11.65 × | 8.47 × |

Table 6: Transition Point & EDP gain for ERGo compared to FpT with different thresholds of acceptable accuracy loss

| | **Δ Acc =1%** | | **Δ Acc =0.5%** | | **Δ Acc =0.1%** | |
|---|---|---|---|---|---|---|
| Dataset | Transition Point (%) | EDP Gain | Transition Point (%) | EDP Gain | Transition Point (%) | EDP Gain |
| PPI | 6 | 17.31× | 11 | 11.09 × | 13 | 9.60 × |
| Reddit | 4 | 15.29 × | 4 | 15.29 × | 5 | 13.95 × |
| Amazon | 3 | 13.53 × | 4 | 12.49 × | 5 | 11.61 × |
| Cora | 6 | 13.52 × | 8 | 11.55 × | 9 | 10.77 × |
| Ogbl | 3 | 24.63 × | 3 | 24.63 × | 5 | 18.68 × |

Moreover, the CP framework enables us to precisely control and specify user-specified accuracy-loss thresholds (such as 0.1% or 1%). This is achieved by training the regression model using an output label corresponding to the acceptable accuracy. Note that ERGo guarantees that the true switching epoch is between the bounds determined through CP with a high probability, say 95%. This enables ERGo to effectively trade-off between accuracy and resource efficiency while GNN training. For instance, the results presented in Figs. 9 to 12 for ERGo utilizes a transition point determined by a regression model trained with an output label of 1% accuracy loss. The resulting drop in accuracy for datasets is between 0.1% and 0.5%, except for PPI, where the accuracy loss is 0.8% (see Fig. 6). In scenarios, where preserving model accuracy is of primary concern, we can train the regression model with a label corresponding to a lower accuracy loss threshold (say 0.1%). Table 6 details this trade-off between accuracy and EDP gain against the baseline FpT under different thresholds of acceptable accuracy loss for various GNN datasets. Now, for PPI, we were able to reduce the accuracy loss to 0.2% by training the regression model using an output label corresponding to a 0.5% accuracy loss, instead of a 1% loss. However, this results in a delayed transition to the parameter-efficient phase and a performance penalty. Specifically, the resultant EDP gain decreases from 17.31× (with a 1% threshold) to 11.09× (with a 0.5% threshold) compared to FpT. Finally, the results in Table 6 also highlight the significant variation in transition points among datasets. For instance, the Ogbl dataset exhibits a transition point at 5% with a 0.1% accuracy loss threshold, whereas the PPI dataset transition point is at 13% under the same threshold. This observation underscores the limitation of fixed arbitrary thresholds, highlighting the need for dynamically determining the transition from a full-parameter phase to a parameter-efficient phase based on the intrinsic characteristics of the GNN training task.

## 6 CONCLUSION

In this paper, we present ERGo, a hybrid training framework for GNN training. ERGo employs a calibrated regression model using conformal prediction to determine the transition point from the full-parameter phase to the parameter-efficient phase during GNN training in a provably safe manner. Within the parameter-efficient phase, ERGo leverages low-rank adaptations to reduce the computational complexity of back-propagation while maintaining GNN model accuracy. Notably, ERGo is compatible with existing PIM architectures employed for GNN training, offering significant enhancements in both performance and energy efficiency. Furthermore, ERGo prevents rewrites onto NVM crossbars in a heterogeneous NVM-SRAM system during the parameter-efficient phase of training, which addresses the endurance issues commonly associated with NVM devices. Experimental results demonstrate that ERGo reduces off-chip access by 5.7× and achieves up to 24.63× reduction in energy-delay product (EDP) compared to full-parameter training on the same hardware platform.

## ACKNOWLEDGMENTS

This work was supported by the US National Science Foundation grant CSR-2308530, by the Army Research Office under Grant ARO-W911NF-24-1-0240 and partially supported by the Intel CAD SRS Program.